\documentclass[11pt,a4paper]{article}
\usepackage[hyperref]{emnlp2018}
\usepackage{times}
\usepackage{latexsym}
\usepackage{url}
\usepackage{scalefnt}
\usepackage{booktabs}
\usepackage{multirow}
\usepackage{graphicx}
\usepackage{microtype}
\usepackage{tabularx}
\usepackage{amsmath}
\usepackage{amssymb}
\usepackage{xspace}
\usepackage{capt-of}

\newcommand{\ch}{\checkmark}
\newcommand{\nh}{}
\newcommand{\F}{\text{F$_1$}\xspace}
\newcommand{\eg}{\textit{e.\,g.}\xspace}
\newcommand{\ie}{\textit{i.\,e.}\xspace}
\newcommand{\ia}{\textit{i.\,a.}\xspace}
\newcommand{\emotionword}[1]{\textit{#1}}
\newcommand{\anger}{\emotionword{anger}}
\newcommand{\joy}{\emotionword{joy}}
\newcommand{\fear}{\emotionword{fear}}
\newcommand{\sadness}{\emotionword{sadness}}
\newcommand{\disgust}{\emotionword{disgust}}
\newcommand{\surprise}{\emotionword{surprise}}
\newcommand{\rt}[1]{\multicolumn{1}{c}{\rotatebox{90}{#1}}}

\newcommand{\ccol}[1]{\multicolumn{1}{c}{#1}}
\aclfinalcopy

\title{IEST: WASSA-2018 Implicit Emotions Shared Task\\[8mm]}

\author{Roman Klinger$^1$, Orph\'ee De Clercq$^2$, Saif M.\ Mohammad$^3$,
  \and Alexandra Balahur$^4$\\[5mm]
  $^1$ Institut f\"ur Maschinelle Sprachverarbeitung, Universit\"at Stuttgart, Germany\\
  \texttt{klinger@ims.uni-stuttgart.de}\\
  $^2$ LT3, Language and Translation Technology Team, Ghent University, Belgium\\
  \texttt{orphee.declercq@ugent.be}\\
  $^3$ National Research Council Canada, Ottawa, Ontario, Canada\\
  \texttt{saif.mohammad@nrc-cnrc.gc.ca}\\
  $^4$ Text and Data Mining Unit, European Commission Joint Research Centre, Ispra, Italy\\
  \texttt{alexandra.balahur@ec.europa.eu}\\
}

\date{}

\begin{document}
\maketitle

\begin{abstract}
  Past shared tasks on emotions use data with both overt expressions
  of emotions ({\it I am so happy to see you!}) as well as subtle
  expressions where the emotions have to be inferred, for instance
  from event descriptions.  Further, most datasets do not focus on the
  cause or the stimulus of the emotion. Here, for the first time, we
  propose a shared task where systems have to predict the emotions in
  a large automatically labeled dataset of tweets without access to
  words denoting emotions. Based on this intention, we call this the
  Implicit Emotion Shared Task (IEST) because the systems have to
  infer the emotion mostly from the context. Every tweet has an
  occurrence of an explicit emotion word that is masked.  The tweets
  are collected in a manner such that they are likely to include a
  description of the cause of the emotion -- the stimulus.
  Altogether, 30 teams submitted results which range from macro \F
  scores of 21\,\% to 71\,\%. The baseline (MaxEnt bag of words and
  bigrams) obtains an \F score of 60\,\% which was available to
  the participants during the development phase.
  A study with human annotators suggests that automatic methods
  outperform human predictions, possibly by honing into subtle textual
  clues not used by humans.
  Corpora, resources, and results are available at the shared task
  website at \url{http://implicitemotions.wassa2018.com}.
\end{abstract}

\section{Introduction}
\scalefont{0.99}
The definition of emotion has long been debated. The main subjects of
discussion are the origin of the emotion (physiological or cognitive),
the components it has (cognition, feeling, behaviour) and the manner
in which it can be measured (categorically or with continuous
dimensions). The Implicit Emotion Shared Task (IEST) is based on
\newcite{Scherer2005}, who considers emotion as ``an episode of
interrelated, synchronized changes in the states of all or most of the
five organismic subsystems (information processing, support,
executive, action, monitor) in response to the evaluation of an
external or internal stimulus event as relevant to major concerns of
the organism''.

This definition suggests that emotion is triggered by
the interpretation of a stimulus event (\ie, a situation) according to
its meaning, the criteria of relevance to the personal goals, needs,
values and the capacity to react. As such, while most situations will
trigger the same emotional reaction in most people, there are
situations that may trigger different affective responses in different
people. This is explained more in detail by the psychological theories
of emotion known as the ``appraisal theories'' \cite{Scherer2005}.

Emotion recognition from text is a research area in natural language
processing (NLP) concerned with the classification of words, phrases, or
documents into predefined emotion categories or dimensions. Most
research focuses on \textit{discrete emotion recognition}, which
assigns categorical emotion labels \cite{Ekman1992,Plutchik2001}, \eg,
\textit{Anger}, \textit{Anticipation}, \textit{Disgust},
\textit{Fear}, \textit{Joy}, \textit{Sadness}, \textit{Surprise} and
\textit{Trust}.\footnote{Some shared tasks on fine emotion intensity
  include the SemEval-2007 Task 14, WASSA-2017 shared task EmoInt
  \cite{MohammadB17wassa}, and SemEval-2018 Task 1
  \cite{SemEval2018Task1}.}  Previous research developed statistical,
dictionary, and rule-based models for several domains, including fairy
tales \cite{Alm2005}, blogs \cite{Aman2007} and microblogs
\cite{Dodds2011}. Presumably, most models built on such datasets rely
on emotion words (or their representations) whenever accessible and are
therefore not forced to learn associations for more subtle
descriptions. Such models might fail to predict the correct emotion
when such overt words are not accessible. Consider the instance ``when
my child was born'' from the ISEAR corpus, a resource in which people
have been asked to report on events when they felt a specific
predefined emotion. This example does not contain any emotion word
itself, though one might argue that the words ``child'' and ``born''
have a positive prior connotation.

\newcite{BalahurDSS2012} showed that the inference of affect from text
often results from the interpretation of the situation presented
therein. Therefore, specific approaches have to be designed to
understand the emotion that is generally triggered by situations. Such
approaches require common sense and world knowledge
\cite{Liu2003,Cambria2009}. Gathering
world knowledge to support NLP is challenging,
although different resources have been built to this
aim -- \eg, Cyc\footnote{http://www.cyc.com} and ConceptNet
\cite{Liu2004}.

On a different research branch, the field of distant supervision and
weak supervision addresses the challenge that manually annotating data
is tedious and expensive. Distant supervision tackles this by making
use of structured resources to automatically label data
\cite{Mintz2009,Riedel2010,mohammad:2012:STARSEM}. This approach has been adapted in emotion
analysis by using information assigned by authors to their own text,
with the use of hashtags and emoticons \cite{Wang2012b}.

%\bigskip\noindent 
With the\textit{ Implicit Emotion Shared Task (IEST)}, we aim at
combining these two research branches: On the one hand, we use distant
supervision to compile a corpus of substantial size. On the other
hand, we limit the corpus to those texts which are likely to contain
descriptions of the cause of the emotion -- the \textit{stimulus}. Due
to the ease of access and the variability and data richness on
Twitter, we opt for compiling a corpus of microposts, from which we
sample tweets that contain an emotion word followed by `that', `when',
or `because'.  We then mask the emotion word and ask systems to
predict the emotion category associated with that word.\footnote{This
  gives the shared task a mixed flavor of both text classification and
  word prediction, in the spirit of distributional semantics.}  The
emotion category can be one of six classes: \anger, \disgust, \fear,
\joy, \sadness, and \surprise.  Examples from the data are:
\begin{quote}
  (1) ``It's [\#TARGETWORD\#] when you feel like you are invisible to others.''\\[2mm]
  (2) ``My step mom got so [\#TARGETWORD\#] when she came home from work and saw
that the boys didn't come to Austin with me.''
\end{quote}
In Example 1, the inference is that feeling invisible typically makes
us sad. In Example 2, the context is presumably that a person (mom)
expected something else than what was expected. This in isolation
might cause anger or sadness, however, since ``the boys are home'' the
mother is likely happy. Note that such examples can be used as source
of commonsense or world knowledge to detect emotions from contexts
where the emotion is not explicitly implied.

The shared task was conducted between 15 March 2018 (publication of
train and trial data) and the evaluation phase, which ran from 2 to 9
July. Submissions were managed on
CodaLab\footnote{\url{https://competitions.codalab.org/competitions/19214}}.
The best performing systems are all ensembles of deep learning
approaches. Several systems make use of external additional resources
such as pre-trained word vectors, affect lexicons, and language models
fine-tuned to the task.

The rest of the paper is organized as follows: we first review related
work (Section~\ref{sec:relatedwork}). Section~\ref{sec:sharedtask}
introduces the shared task, the data used, and the setup. The results
are presented in Section~\ref{sec:results}, including the official
results and a discussion of different submissions. The automatic
system's predictions are then compared to human performance in
Section~\ref{sec:human}, where we report on a crowdsourcing study with
the data used for the shared task. We conclude in
Section~\ref{sec:conclusion}.

\section{Related Work}
\label{sec:relatedwork}
Related work is found in different directions of research on emotion
detection in NLP: resource creation and emotion classification, as
well as modeling the emotion itself.

Modeling the emotion computationally has been approached from the
perspective of humans needs and desires with the goal of simulating human
reactions. \newcite{Dyer87} presents three models which take into
account characters, arguments, emotion experiencers, and events. These
aspects are modeled with first order logic in a procedural
manner. Similarly, \newcite{SubasicHuettner2001} use fuzzy logic for
such modeling in order to consider gradual differences.  A similar
approach is followed by the OCC model \cite{Ortony1990}, for which
\newcite{Udochukwu2015} show how to connect it to text in a rule-based
manner for implicit emotion detection. Despite of this early work on
holistic computational models of emotions, NLP
focused mostly on a more coarse-grained level.

One of the first corpora annotated for emotions is that by
\newcite{Alm2005} who analyze sentences from fairy tales.
\newcite{Strapparava2007} annotate news headlines with emotions and
valence, \newcite{Mohammad2015a} annotate tweets on elections, and
\newcite{Schuff2017} tweets of a stance dataset \citep{Mohammad2017c}.
The SemEval-2018 Task 1: Affect in Tweets \cite{SemEval2018Task1}
includes several subtasks on inferring the affectual state of a person
from their tweet: emotion intensity regression, emotion intensity
ordinal classification, valence (sentiment) regression, valence
ordinal classification, and multi-label emotion classification.  In
all of these prior shared tasks and datasets, no distinction is made
between implicit or explicit mentions of the emotions. We refer the
reader to \newcite{Bostan2018} for a more detailed overview of emotion
classification datasets.

Few authors specifically analyze which phrase triggers the perception
of an emotion.  \newcite{Aman2007} focus on the annotation on document
level but also mark emotion indicators.  \newcite{Mohammad2014}
annotate electoral tweets for semantic roles such as emotion and
stimulus (from FrameNet).  \newcite{Ghazi2015} annotate a subset of
\newcite{Aman2007} with causes (inspired by the FrameNet
structure).  \newcite{Kim2018} and
\newcite{neviarouskaya2013extracting} similarly annotate emotion
holders, targets, and causes as well as the trigger words.

One of the oldest resources nowadays used for emotion recognition is
the ISEAR set \citep{Scherer1997} which consists of self-reports of
emotional events. As the task of participants in a psychological study was not to 
express an emotion but to report on an event in which they experienced
a given emotion, this resource can be considered similar to our goal
of focusing on implicit emotion expressions.

With the aim to extend the coverage of ISEAR,
\newcite{Balahur2011,Balahur2012} build EmotiNet, a knowledge base to
store situations and the affective reactions they have the potential
to trigger. They show how the knowledge stored can be expanded using
lexical and semantic similarity, as well as through the use of
Web-extracted knowledge \cite{Balahur2013}. The patterns used to
populate the database are of the type ``I feel [emotion] when
[situation]'', which was also a starting point for our task.

Finally, several approaches take into consideration distant
supervision
\citep[\ia]{Mohammad2015b,Abdul2017,Choudhury2012,Liu2017}. This is
motivated by the high availability of user-generated text and by the
challenge that manual annotation is typically tedious or
expensive. This contrasts with the current data demand of machine
learning, and especially, deep learning approaches.

With our work in IEST, we combine the goal of the development of
models which are able to recognize emotions from implicit descriptions
without having access to explicit emotion words, with the paradigm of
distant supervision.

\section{Shared Task}
\label{sec:sharedtask}
\subsection{Data}
The aim of the Implicit Emotion Shared Task is to force models to
infer emotions from the context of emotion words without having access
to them. Specifically, the aim is that models infer the emotion
through the causes mentioned in the text.  Thus, we build the corpus
of Twitter posts by polling the Twitter
API\footnote{\url{https://developer.twitter.com/en/docs.html}} for the
expression `\verb!EMOTION-WORD (that|because|when)!', where
\verb!EMOTION-WORD! contains a synonym for one out of six
emotions.\footnote{Note that we do not check that there is a white
  space before the emotion word, which leads to tweets containing
  \ldots ``unEMOTION-word\ldots''.} The synonyms are shown in Table
\ref{tab:syns}.  The requirement of tweets to have either `that',
`because', or `when' immediately after the emotion word means that the
tweet likely describes the cause of the emotion.

\begin{table}
  \centering\small
    \begin{tabularx}{\linewidth}{llX}
      \toprule
      Emotion & Abbr. & Synonyms \\
      \cmidrule(r){1-1}\cmidrule(lr){2-2}\cmidrule(l){3-3}
      Anger & A & angry, furious\\
      Fear & F & afraid, frightened, scared, fearful\\
      Disgust & D & disgusted, disgusting\\
      Joy & J & cheerful, happy, joyful\\
      Sadness  & Sa & sad, depressed, sorrowful\\
      \multirow{2}{*}{Surprise} & \multirow{2}{*}{Su} & surprising, surprised, astonished, shocked, startled, astounded, stunned\\
      \bottomrule
    \end{tabularx}
  \caption{Emotion synonyms used when polling Twitter.}
  \label{tab:syns}
\end{table}

The initially retrieved large dataset has a distribution of 25\,\%
\surprise, 23\,\% \sadness, 18\,\% \joy, 16\,\% \fear, 10\,\% \anger,
8\,\% \disgust.  We discard tweets with more than one emotion word, as
well as exact duplicates, and mask usernames and URLs. From this set,
we randomly sample 80\,\% of the tweets to form the training set
(153,600 instances), 5\,\% as trial set (9,600 instances), and 15\,\%
as test set (28,800 instances). We perform stratified sampling to
obtain a balanced dataset.  While the shared task took place, two
errors in the data preprocessing were discovered by participants (the
use of the word unhappy as synonym for sadness, which lead to
inconsistent preprocessing in the context of negated expressions, and
the occurrence of instances without emotion words). To keep the change
of the data at a minimum, the erroneous instances were only removed,
which leads to a distribution of the data as shown in
Table~\ref{tab:distribution}.

% weka:
% k=0 ; cat trial.csv | while read i ; do folder=`echo $i | cut -f 1` ; mkdir -p $folder ; echo $i | cut -f 2 > $folder/$k.txt ; k=$(($k+1)) ; done
% java -cp ~/opt/weka/weka-3-8-1/weka.jar weka.core.converters.TextDirectoryLoader -dir . > trial.arff

\begin{table}
  \centering\small
  \begin{tabular}{lrrr}
    \toprule
    Emotion & \ccol{Train} & \ccol{Trial} & \ccol{Test} \\
    \cmidrule(r){1-1}\cmidrule(rl){2-2}\cmidrule(rl){3-3}\cmidrule(l){4-4}
        Anger    & 25562  & 1600 & 4794 \\
   Disgust  & 25558  & 1597 & 4794 \\
    Fear     & 25575  & 1598 & 4791 \\
     Joy      & 27958  & 1736 & 5246 \\
     Sadness  & 23165  & 1460 & 4340 \\
    Surprise & 25565  & 1600 & 4792 \\

    \cmidrule(r){1-1}\cmidrule(rl){2-2}\cmidrule(rl){3-3}\cmidrule(l){4-4}
    Sum      & 153383 & 9591 & 28757 \\
    \bottomrule
  \end{tabular}
  \caption{Distribution of IEST data.}
  \label{tab:distribution}
\end{table}

\subsection{Task Setup}
The shared task was announced through a dedicated website
(\url{http://implicitemotions.wassa2018.com/}) and
computational-linguistics-specific mailing lists. The organizers
published an evaluation script which calculates precision, recall, and
\F~measure for each emotion class as well as micro and macro
average. Due to the nearly balanced dataset, the chosen official
metric for ranking submitted systems is the macro-\F measure.

In addition to the data, the participants were provided a list of
resources they might want to
use\footnote{\url{http://implicitemotions.wassa2018.com/resources/}}
(and they were allowed to use any other resources they have access to
or create themselves).  We also provided access to a baseline
system.\footnote{\url{https://bitbucket.org/rklinger/simpletextclassifier}}
This baseline is a maximum entropy classifier with L2
regularization. Strings which match \verb![#a-zA-Z0-9_=]+|[^ ]! form
tokens. As preprocessing, all symbols which are not alphanumeric or
contain the \# sign are removed. Based on that, unigrams and bigrams
form the Boolean features as a set of words for the classifier.

\section{Results}
\label{sec:results}
\subsection{Baseline}
The intention of the baseline implementation was to provide
participants with an intuition of the difficulty of the task. It
reaches 59.88\,\% macro \F on the test data, which is very similar to
the trial data result (60.1\,\% \F). The confusion matrix for the
baseline is presented in Table~\ref{tab:confusionbaseline}; the
confusion matrix for the best submitted system is shown in
Table~\ref{tab:confusionbest}.

\begin{table}
  \centering\small
  \begin{tabular}{l|lrrrrrr}
    \toprule
    \multicolumn{2}{c}{}&\multicolumn{6}{c}{Predicted Labels}\\
    \cmidrule(l){3-8}
    \multicolumn{2}{c}{}& A & D & F & J & Sa & Su \\
    \cmidrule(r){3-3}\cmidrule(rl){4-4}\cmidrule(rl){5-5}\cmidrule(rl){6-6}\cmidrule(rl){7-7}\cmidrule(rl){8-8}
    \multirow{6}{*}{\rotatebox{90}{Gold Labels}}
    & A  & 2431 & 476 & 496 & 390 & 410 & 426 \\
    & D  & 426 & 2991 & 245 & 213 & 397 & 522 \\
    & F  & 430 & 249 & 3016 & 327 & 251 & 518 \\
    & J  & 378 & 169 & 290 & 3698 & 366 & 345 \\
    & Sa & 450 & 455 & 313 & 458 & 2335 & 329 \\
    & Su & 411 & 508 & 454 & 310 & 279 & 2930 \\
    \bottomrule
  \end{tabular}
  \caption{Confusion Matrix on Test Data for Baseline.}
  \label{tab:confusionbaseline}
\end{table}

\begin{table}
  \centering\small
  \begin{tabular}{l|lrrrrrr}
    \toprule
    \multicolumn{2}{c}{}&\multicolumn{6}{c}{Predicted Labels}\\
    \cmidrule(l){3-8}
    \multicolumn{2}{c}{}& A & D & F & J & Sa & Su \\
    \cmidrule(r){3-3}\cmidrule(rl){4-4}\cmidrule(rl){5-5}\cmidrule(rl){6-6}\cmidrule(rl){7-7}\cmidrule(rl){8-8}
    \multirow{6}{*}{\rotatebox{90}{Gold Labels}}
    & A & 3182 & 313 & 293 & 224 & 329 & 453 \\
    & D & 407 & 3344 & 134 & 102 & 336 & 471 \\
    & F & 403 & 129 &  3490 & 196 & 190 & 383 \\
    & J & 297 & 67 & 161 & 4284 & 220 & 217 \\
    & Sa& 443 & 340 & 171 & 240 & 2947 & 199 \\
    & Su& 411 & 367 & 293 & 209 & 176 & 3336 \\
    \bottomrule
  \end{tabular}
  \caption{Confusion Matrix on Test Data of Best Submitted System}
  \label{tab:confusionbest}
\end{table}

% fr: feedback on tables received
\begin{table}[t]
\centering\small\scalefont{1.0}
\setlength\tabcolsep{8pt}
\newcommand{\sep}{\cmidrule(r){1-1}\cmidrule(r){2-2}\cmidrule(r){3-3}\cmidrule(rl){4-4}\cmidrule(l){5-5}}
\begin{tabular}{llccr}
\toprule
id & Team & \F & Rank & B \\
\sep
1&\textit{Amobee}&\textit{71.45}&\textit{(1)}& 3 \\ % fr 
2&IIIDYT&71.05&(2) & 3 \\ % fr
3&NTUA-SLP&70.29&(3) & 4  \\ % fr
4&UBC-NLP&69.28&(4) & 6 \\ % fr
5&Sentylic&69.20&(5) & 7  \\
6&HUMIR&68.64&(6)& 8 \\ % fr
7&nlp&68.48&(7) & 9 \\
8&DataSEARCH&68.04&(8) & 10 \\
9&YNU1510&67.63&(9)& 11 \\ % fr
10&EmotiKLUE&67.13&(10) & 11 \\
11&wojtek.pierre&66.15&(11) & 15 \\
12&hgsgnlp&65.80&(12)& 15 \\
13&UWB&65.70&(13) & 15 \\
14&NL-FIIT&65.52&(14) & 15 \\
15&TubOslo&64.63&(15) & 17 \\
16&YNU\_Lab&64.10&(16) & 17 \\ % fr
17&Braint&62.61&(17) & 19 \\
18&EmoNLP&62.11&(18) & 19 \\
19&RW&60.97&(19) & 20 \\
\sep
20&Baseline&59.88&& 21 \\
\sep
21&USI-IR&58.37&(20) & 22 \\
22&THU\_NGN&58.01&(21) & 23 \\
23&SINAI&57.94&(22) & 24 \\
24&UTFPR&56.92&(23) & 26 \\
25&\textcolor{gray}{CNHZ2017}&56.40&& 27 \\
26&\textcolor{gray}{lyb3b}&55.87&& 27 \\
27&Adobe Research&53.08&(24) & 28 \\
28&\textcolor{gray}{Anonymous}&50.38& & 29 \\ % ColumbiaNLP
29&dinel&49.99&(25) & 30 \\
30&CHANDA&41.89&(26) & 31 \\
31&\textcolor{gray}{NLP\_LDW}&21.03&& \\
\bottomrule
\end{tabular}
\caption{Official results of IEST 2018. Participants who did not
  report on the system details did not get assigned a rank and are
  reported in gray. Column B provides the
  first row in the results table to which the respective row is
  significantly different (confidence level 0.99), tested with
  bootstrap resampling.
}
\label{tab:offresults}
\end{table}

\subsection{Submission Results}
Table~\ref{tab:offresults} shows the main results of the shared
task. We received submissions through CodaLab from thirty
participants. Twenty-six teams responded to a post-competition survey providing additional information regarding
team members (56 people in total) and the systems that were
developed. For the remaining analyses and the ranking, we only report
on these twenty-six teams.

The table shows results from 31 systems, including the baseline
results which have been made available to participants during the
shared task started. From all submissions, 19 submissions scored above
the baseline. The best scoring system is from team \textit{Amobee},
followed by \textit{IIDYT} and \textit{NTUA-SLP}. The first two
results are not significantly different, as tested with the \newcite{Wilcoxon1945} sign test ($p<0.01$) and with bootstrap resampling (confidence level 0.99).

%ODC: do we add Table 4 somewhere "Confusion matrix of the Best Submitted System"? Now this is not discussed throughout the paper.

Table \ref{tab:emotresults} in the Appendix shows a breakdown of the
results by emotion class. Though the data was nearly balanced, joy is
mostly predicted with highest performance, followed by fear and
disgust. The prediction of surprise and anger shows a lower
performance.

Note that the macro \F evaluation took into account all classes which
were either predicted or in the gold data. Two teams submitted results
which contain labels not present in the gold data, which reduced the
macro-\F dramatically. With an evaluation only taking into account 6
labels, id 22 would be on rank 9 and id 28 would be on rank 10.

\begin{table}[t]
  \centering\small
  \setlength\tabcolsep{1mm}
  \renewcommand*{\arraystretch}{0.7}
  \begin{tabular}{r|rrrrrrrrrrrrrrr}
    \toprule
   \rt{Rank} &\rt{Keras}&\rt{Tensorflow}&\rt{Pandas}&\rt{SciKitLearn}&\rt{NLTK}&\rt{GloVe}&\rt{Gensim}&\rt{PyTorch}&\rt{FastText}&\rt{SpaCy}&\rt{Weka}&\rt{ELMo}&\rt{LibLinear}&\rt{Theano} \\
\cmidrule(r){1-1}\cmidrule(l){2-15}
1&\ch&\ch&\ch&\nh&\nh&\nh&\ch&\nh&\nh&\nh&\nh&\nh&\nh&\nh\\ % fr
2&\nh&\nh&\nh&\nh&\nh&\nh&\nh&\ch&\nh&\nh&\nh&\ch&\nh&\nh\\ % fr
3&\nh&\nh&\ch&\ch&\nh&\nh&\nh&\ch&\nh&\nh&\nh&\nh&\nh&\nh\\ % fr
4&\ch&\nh&\ch&\ch&\nh&\nh&\nh&\ch&\ch&\ch&\nh&\ch&\nh&\nh\\ % fr
5&\ch&\ch&\ch&\ch&\nh&\nh&\nh&\nh&\nh&\nh&\nh&\nh&\nh&\ch\\
6&\ch&\ch&\nh&\nh&\ch&\ch&\nh&\nh&\nh&\nh&\ch&\nh&\nh&\nh\\ % fr
7&\ch&\ch&\nh&\ch&\ch&\ch&\ch&\nh&\nh&\nh&\nh&\nh&\nh&\nh\\
8&\ch&\ch&\ch&\ch&\ch&\ch&\ch&\nh&\ch&\nh&\nh&\nh&\nh&\nh\\
9&\ch&\ch&\ch&\ch&\ch&\ch&\ch&\nh&\ch&\nh&\nh&\nh&\nh&\nh\\ % fr
10&\ch&\nh&\nh&\nh&\nh&\nh&\ch&\nh&\nh&\nh&\nh&\nh&\nh&\nh\\
11&\nh&\nh&\ch&\nh&\nh&\nh&\ch&\ch&\ch&\ch&\nh&\ch&\nh&\nh\\
12&\ch&\nh&\ch&\nh&\nh&\ch&\ch&\nh&\ch&\nh&\nh&\nh&\nh&\nh\\
13&\ch&\ch&\ch&\ch&\nh&\nh&\ch&\nh&\nh&\nh&\ch&\nh&\nh&\nh\\
14&\ch&\ch&\ch&\nh&\ch&\ch&\nh&\nh&\nh&\nh&\nh&\nh&\nh&\nh\\
15&\nh&\nh&\nh&\ch&\nh&\nh&\nh&\nh&\nh&\nh&\nh&\nh&\nh&\nh\\
16&\ch&\ch&\ch&\nh&\nh&\ch&\ch&\nh&\nh&\nh&\nh&\nh&\nh&\nh\\ % fr
17&\ch&\nh&\nh&\nh&\ch&\nh&\nh&\nh&\nh&\nh&\nh&\nh&\nh&\nh\\
18&\ch&\ch&\nh&\ch&\nh&\ch&\nh&\nh&\nh&\nh&\nh&\nh&\nh&\nh\\
19&\nh&\nh&\nh&\nh&\ch&\nh&\nh&\nh&\nh&\nh&\nh&\nh&\ch&\nh\\
20&\nh&\ch&\ch&\ch&\ch&\nh&\ch&\nh&\nh&\nh&\nh&\nh&\nh&\nh\\
21&\ch&\ch&\nh&\ch&\ch&\nh&\nh&\nh&\nh&\nh&\ch&\nh&\nh&\nh\\
22&\ch&\ch&\ch&\nh&\ch&\ch&\nh&\nh&\nh&\nh&\nh&\nh&\nh&\nh\\
23&\nh&\nh&\nh&\nh&\nh&\nh&\ch&\ch&\nh&\nh&\nh&\nh&\nh&\nh\\
24&\nh&\ch&\ch&\ch&\ch&\ch&\nh&\nh&\nh&\ch&\nh&\nh&\nh&\nh\\
25&\nh&\nh&\nh&\ch&\ch&\ch&\nh&\nh&\nh&\nh&\nh&\nh&\nh&\nh\\
26&\nh&\nh&\nh&\nh&\ch&\nh&\nh&\nh&\nh&\nh&\nh&\nh&\nh&\nh\\
\cmidrule(r){1-1}\cmidrule(l){2-15}
&16&14&14&13&13&11&11&5&5&3&3&3&1&1\\
\bottomrule
\end{tabular}
\caption{Overview of tools employed by different teams (sorted by
  popularity from left to right).}
\label{tab:toolsoverview}
\end{table}

\begin{table}[t]
  \centering\small
  \setlength\tabcolsep{0.85mm}
  \renewcommand*{\arraystretch}{0.7}
  \begin{tabular}{r|rrrrrrrrrrrrrrr}
    \toprule
\rt{Rank}&\rt{Embeddings}&\rt{LSTM/RNN/GRU}&\rt{Ensemble}&\rt{CNN/Capsules}&\rt{Attention}&\rt{Linear
  Classifier}&\rt{Transfer Learning}&\rt{Language
  model}&\rt{MLP}&\rt{Autoencoder}&\rt{Random
  Forrest}&\rt{k-Means}&\rt{Bagging}&\rt{LDA}\\
\cmidrule(r){1-1}\cmidrule(l){2-15}
1&\ch&\ch&\ch&\ch&\ch&\nh&\ch&\ch&\nh&\nh&\nh&\nh&\nh&\nh\\ % fr
2&\ch&\ch&\ch&\nh&\nh&\nh&\nh&\nh&\ch&\nh&\nh&\nh&\nh&\nh\\ % fr
3&\ch&\ch&\ch&\nh&\ch&\nh&\ch&\ch&\nh&\nh&\nh&\nh&\nh&\nh\\ % fr
4&\ch&\ch&\ch&\nh&\ch&\nh&\ch&\ch&\nh&\nh&\nh&\nh&\nh&\nh\\
5&\ch&\ch&\ch&\ch&\nh&\nh&\nh&\nh&\nh&\nh&\nh&\nh&\nh&\nh\\
6&\ch&\ch&\ch&\nh&\nh&\nh&\ch&\nh&\ch&\nh&\nh&\nh&\ch&\nh\\ % fr
7&\ch&\ch&\ch&\nh&\ch&\nh&\nh&\nh&\nh&\nh&\nh&\nh&\nh&\nh\\
8&\ch&\ch&\nh&\ch&\nh&\nh&\nh&\nh&\nh&\nh&\nh&\nh&\nh&\nh\\
9&\ch&\ch&\ch&\ch&\ch&\nh&\nh&\nh&\nh&\nh&\nh&\nh&\nh&\nh\\ % fr
10&\ch&\ch&\nh&\nh&\nh&\nh&\nh&\nh&\nh&\nh&\nh&\nh&\nh&\ch\\
11&\ch&\ch&\nh&\ch&\nh&\nh&\nh&\nh&\nh&\nh&\nh&\nh&\nh&\nh\\
12&\ch&\nh&\ch&\ch&\nh&\nh&\nh&\nh&\nh&\nh&\nh&\ch&\nh&\nh\\
13&\ch&\ch&\nh&\nh&\nh&\nh&\nh&\nh&\nh&\nh&\nh&\nh&\nh&\nh\\
14&\ch&\ch&\nh&\nh&\nh&\nh&\ch&\nh&\nh&\nh&\nh&\nh&\nh&\nh\\
15&\nh&\nh&\nh&\nh&\nh&\ch&\nh&\nh&\nh&\nh&\nh&\nh&\nh&\nh\\
16&\ch&\ch&\nh&\nh&\ch&\nh&\nh&\nh&\nh&\nh&\nh&\nh&\nh&\nh\\ % fr
17&\ch&\ch&\nh&\nh&\ch&\ch&\nh&\nh&\nh&\nh&\nh&\nh&\nh&\nh\\
18&\ch&\ch&\ch&\ch&\nh&\nh&\nh&\nh&\nh&\nh&\nh&\nh&\nh&\nh\\
19&\nh&\nh&\nh&\nh&\nh&\ch&\nh&\nh&\nh&\nh&\nh&\nh&\nh&\nh\\
20&\ch&\ch&\nh&\nh&\nh&\nh&\nh&\nh&\nh&\nh&\nh&\nh&\nh&\nh\\
21&\ch&\ch&\ch&\ch&\nh&\ch&\nh&\nh&\nh&\nh&\nh&\nh&\nh&\nh\\
22&\ch&\ch&\nh&\nh&\nh&\nh&\nh&\nh&\nh&\nh&\nh&\nh&\nh&\nh\\
23&\ch&\ch&\nh&\nh&\nh&\nh&\nh&\nh&\nh&\nh&\nh&\nh&\nh&\nh\\
24&\ch&\nh&\ch&\ch&\nh&\nh&\nh&\nh&\nh&\ch&\nh&\nh&\nh&\nh\\
25&\ch&\nh&\nh&\nh&\nh&\ch&\nh&\nh&\nh&\nh&\ch&\nh&\nh&\nh\\
26&\nh&\nh&\nh&\nh&\nh&\nh&\nh&\nh&\nh&\nh&\nh&\nh&\nh&\nh\\
\cmidrule(r){1-1}\cmidrule(l){2-15}
&23&20&12&9&7&5&5&3&2&1&1&1&1&1\\
\bottomrule
\end{tabular}
\caption{Overview of methods employed by different teams (sorted by
  popularity from left to right).}
\label{tab:methodsoverview}
\end{table}

\begin{table}[t]
  \centering\small
  \setlength\tabcolsep{1.9mm}
  \renewcommand*{\arraystretch}{0.7}
  \begin{tabular}{r|rrrrrrrrr}
    \toprule
\rt{Rank}&\rt{Words}&\rt{Lexicons}&\rt{Characters}&\rt{Emoji}&\rt{Unlabeled
  Corpora}&\rt{Emotion Emb.}&\rt{Sentence/Document}&\rt{SemEval}\\
\cmidrule(r){1-1}\cmidrule(l){2-9}
1&\ch&\ch&\nh&\nh&\ch&\ch&\ch&\ch\\%fr
2&\ch&\nh&\ch&\ch&\nh&\nh&\nh&\nh\\%fr
3&\ch&\nh&\nh&\nh&\ch&\nh&\ch&\ch\\%fr
4&\ch&\nh&\nh&\nh&\ch&\nh&\nh&\nh\\
5&\ch&\nh&\nh&\nh&\nh&\nh&\nh&\nh\\
6&\ch&\ch&\nh&\nh&\nh&\ch&\nh&\nh\\%fr
7&\ch&\nh&\nh&\nh&\nh&\nh&\nh&\nh\\
8&\ch&\nh&\ch&\ch&\nh&\nh&\nh&\nh\\
9&\ch&\nh&\nh&\nh&\nh&\nh&\nh&\nh\\%fr
10&\ch&\nh&\nh&\nh&\ch&\nh&\ch&\nh\\
11&\ch&\ch&\ch&\nh&\nh&\nh&\nh&\nh\\
12&\ch&\ch&\ch&\nh&\ch&\nh&\nh&\nh\\
13&\ch&\nh&\nh&\ch&\nh&\nh&\nh&\nh\\
14&\ch&\nh&\nh&\ch&\nh&\nh&\ch&\nh\\
15&\ch&\nh&\ch&\nh&\nh&\nh&\nh&\nh\\
16&\ch&\nh&\nh&\nh&\nh&\nh&\nh&\nh\\%fr
17&\ch&\nh&\nh&\ch&\nh&\nh&\nh&\nh\\
18&\ch&\ch&\ch&\ch&\ch&\ch&\nh&\nh\\
19&\ch&\nh&\nh&\nh&\nh&\nh&\nh&\nh\\
20&\ch&\ch&\nh&\ch&\nh&\nh&\nh&\nh\\
21&\ch&\ch&\ch&\nh&\nh&\nh&\nh&\nh\\
22&\ch&\ch&\nh&\nh&\nh&\nh&\nh&\nh\\
23&\ch&\nh&\ch&\nh&\nh&\nh&\nh&\nh\\
24&\ch&\ch&\nh&\nh&\nh&\ch&\nh&\ch\\
25&\ch&\nh&\nh&\nh&\nh&\nh&\nh&\nh\\
26&\ch&\nh&\nh&\nh&\nh&\nh&\nh&\nh\\
\cmidrule(r){1-1}\cmidrule(l){2-9}
&26&9&8&7&6&4&4&3\\
\bottomrule
\end{tabular}
\caption{Overview of information sources employed by different teams (sorted by
  popularity from left to right).}
\label{tab:informationoverview}
\end{table}

\subsection{Review of Methods}
Table~\ref{tab:toolsoverview} shows that many participants use
high-level libraries like Keras or NLTK. Tensorflow is only of medium
popularity and Theano is only used by one participant.
Table~\ref{tab:methodsoverview} shows a summary of machine learning
methods used by the teams, as reported by themselves.  Nearly
every team uses embeddings and neural networks; many teams use an
ensemble of architectures. Several teams use language models showing a
current trend in NLP to fine-tune those to
specific tasks \cite{Howard2018}. Presumably, those are specifically helpful in
our task due to its word-prediction aspect.

Finally, Table~\ref{tab:informationoverview} summarizes the different
kinds of information sources taken into account by the teams. Several
teams use affect lexicons in addition to word information and
emoji-specific information. The incorporation of statistical knowledge
from unlabeled corpora is also popular.

\subsection{Top 3 Submissions}
In the following, we briefly summarize the approaches used by the top three
teams: Amobee, IIIDYT, and NTUA-SLP. For more information on these
approaches and those of the other teams, we refer the reader to the
individual system description papers. The three best performing
systems are all ensemble approaches. However, they make use of
different underlying machine learning architectures and rely on different kinds of
information.

\subsubsection{Amobee}
The top-ranking system, Amobee, is an ensemble approach of several
models \cite{Rozental2018}. First, the team trains a Twitter-specific
language model based on the transformer decoder architecture using 5B
tweets as training data. This model is used to find the probabilities
of potential missing words, conditional upon the missing word
describing one of the six emotions.  Next, the team applies transfer
learning from the trained models they developed for SemEval 2018 Task
1: Affect in Tweets 
% citation change by request by Amobee team
\cite{Rozental2018b}. Finally, they directly train on the data
provided in the shared task while incorporating outputs from DeepMoji
\cite{Felbo2017} and ``Universal Sentence Encoder'' \cite{Cer2018} as
features.

\subsubsection{IIIDYT}
The second-ranking system, IIIDYT \cite{Balazs2018}, preprocesses the
dataset by tokenizing the sentences (including emojis), and
normalizing the USERNAME, NEWLINE, URL and TRIGGERWORD
indicators. Then, it feeds word-level representations returned by a
pretrained ELMo layer into a Bi-LSTM with 1 layer of 2048 hidden units
for each direction. The Bi-LSTM output word representations are
max-pooled to generate sentence-level representations, followed by a
single hidden layer of 512 units and output size of 6. The team
trains six models with different random initializations, obtains the
probability distributions for each example, and then averages these to
obtain the final label prediction.

\subsubsection{NTUA-SLP}
The NTUA-SLP system \cite{Chronopoulou2018} is an ensemble of three
different generic models. For the first model, the team pretrains
Twitter embeddings with the word2vec skip-gram model using a large
Twitter corpus. Then, these pretrained embeddings are fed to a neural
classifier with 2 layers, each consisting of 400 bi-LSTM units with
attention. For the second model, they use transfer learning of a
pretrained classifier on a 3-class sentiment classification task
(Semeval17 Task4A) and then apply fine-tuning to the IEST
dataset. Finally, for the third model the team uses transfer learning
of a pretrained language model, according to
\newcite{Howard2018}. They first train 3 language models on 3
different Twitter corpora (2M, 3M, 5M) and then they fine-tune them to
the IEST dataset with gradual unfreezing.

\subsection{Error Analysis}
\label{sec:erroranalysis}
Table~\ref{tab:allwrongallcorrect} in the Appendix shows a subsample
of instances which are predicted correctly by all teams (marked as
$+$, including the baseline system and those who did not report on
system details) and that were not predicted correctly by any team
(marked as $-$), separated by correct emotion label.

For the positive examples which are correctly predicted by all teams,
specific patterns reoccur. For \textit{anger}, the author of the first
example encourages the reader not to be afraid~--~a prompt which
might be less likely for other emotions. For several emotions,
single words or phrases are presumably associated with such
emotions, \eg, ``hungry'' with anger, ``underwear'', ``sweat'',
``ewww'' with disgust, ``leaving'', ``depression'' for sadness, ``why
am i not'' for surprise.

Several examples which are all correctly predicted by all teams for
\textit{joy} include the syllable ``un'' preceding the triggerword -- a
pattern more frequent for this emotion than for others.  Another
pattern is the phrase ``fast and furious'' (with furious for
\textit{anger}) which should be considered a mistake in the sampling
procedure, as it refers to a movie instead of an emotion expression.

Negative examples appear to be reasonable when the emotion is given
but may also be valid with other labels than the gold. For \textit{disgust},
respective emotion synonyms are often used as a strong expression
actually referring to other negative emotions. Especially for
\textit{sadness}, the negative examples include comparably long event
descriptions.

\section{Comparison to Human Performance}
\label{sec:human}

%ODC: do we include this table somewhere in this Section? Now, it is not mentioned throughout the paper...
\begin{table}
  \centering\small
  \begin{tabular}{l|lrrrrrr}
    \toprule
    \multicolumn{2}{c}{}&\multicolumn{6}{c}{Predicted Labels}\\
    \cmidrule(l){3-8}
    \multicolumn{2}{c}{}& A & D & F & J & Sa & Su \\
    \cmidrule(r){3-3}\cmidrule(rl){4-4}\cmidrule(rl){5-5}\cmidrule(rl){6-6}\cmidrule(rl){7-7}\cmidrule(rl){8-8}
    \multirow{6}{*}{\rotatebox{90}{Gold Labels}}
    & A & 349 & 40 & 34 & 55 & 95 & 43   \\
    & D & 195 & 92 & 30 & 84 & 157 & 69  \\
    & F & 94 & 20 & 265 & 92 & 120 & 42  \\
    & J & 39 & 6 & 22 & 398 & 36 & 13    \\
    & Sa& 88 & 37 & 23 & 89 & 401 & 46   \\
    & Su& 123 & 25 & 29 & 132 & 53 & 183 \\
    \bottomrule
  \end{tabular}
  \caption{Confusion Matrix Sample Annotated by Humans in Crowdsourcing}
  \label{tab:confusionhuman}
\end{table}
An interesting research question is how accurately native speakers of
a language can predict the emotion class when the emotion word is
removed from a tweet.  Thus we conducted a crowdsourced study asking
humans to perform the same task as proposed for automatic systems in
this shared task.

We sampled 900 instances from the IEST data: 50 tweets for each of the
six emotions in 18 pair-wise combinations with `because', `that', and
`when'.  The tweets and annotation questionnaires were uploaded on a
crowdsourcing platform, Figure Eight (earlier called
CrowdFlower).\footnote{https://www.figure-eight.com} The questionnaire
asked for the best guess for the emotion (Q1) as well as any other
emotion that they think might apply (Q2).

About 5\,\% of the tweets were annotated internally beforehand for Q1
(by one of the authors of this paper). These tweets are referred to as
gold tweets.  The gold tweets were interspersed with other tweets.  If
a crowd-worker got a gold tweet question wrong, they were immediately
notified of the error.  If the worker's accuracy on the gold tweet
questions fell below 70\,\%, they were refused further annotation, and
all of their annotations were discarded. This served as a mechanism to
avoid malicious annotations.

Each tweet is annotated by at least three people.  A total of 3,619
human judgments of emotion associated with the trigger word were
obtained. Each judgment included the best guess for the emotion
(response to Q1) as well as any other emotion that they think might
apply (response to Q2). The answer to Q1 corresponds to the shared
task setting. However, automatic systems were not given the option of
providing additional emotions that might apply (Q2).

The macro \F for predicting the emotion is 45\,\% (Q1, micro \F of
0.47). Observe that human performance is lower than what automatic
systems reach in the shared task. The correct emotion was present in
the top two guessed emotions in 57\,\% of the cases. Perhaps, the
automatic systems are honing in to some subtle systematic regularities
in hope that particular emotion words are used (for example, the function
words in the immediate neighborhood of the target word). It should
also be noted, however, that the data used for human annotations was
only a subsample of the IEST data.

An analysis of subsets of Tweets containing the words
\textit{because}, \textit{that}, and \textit{when} after the emotion
word shows that Tweets with ``that'' are more difficult (41\,\%
accuracy) than with ``when'' (49\,\%) and ``because'' (51\,\%). This
relationship between performance and query string is not observed in
the baseline system -- here, accuracy on the test data (on the data
used for human evaluation) for the ``that'' subset is 61\,\% (60\,\%),
for ``when'' 62\,\% (53\,\%), and for ``because'' 55\,\% (50\,\%) --
therefore, the automatic system is most challenged by ``because'',
while humans are more challenged by ``that''. Please note that this
comparison on the test data is somewhat unfair since for the human
analysis, the data was sampled in a stratified manner, but not for the
automatic prediction.  The test data contains 5635 ``because'' tweets,
13649 with ``that'' and 9474 with ``when''.

There are differences in the difficulty of the task for different
emotions: The accuracy (\F) by emotion is 57\,\% (46\,\%) for anger,
15\,\% (21\,\%) for disgust, 42\,\% (51\,\%) for fear, 77\,\% (58\,\%)
for joy, 59\,\% (52\,\%) for sadness and 34\,\% (39\,\%) for
surprise. The confusion matrix is depicted in
Table~\ref{tab:confusionhuman}. Disgust is often confused with anger,
followed by fear being confused with sadness. Surprise is often
confused with anger and joy.

\section{Conclusions \& Future Work}
\label{sec:conclusion}
With this paper and the Implicit Emotion Shared Task, we presented the
first dataset and joint effort to focus on causal descriptions to
infer emotions that are triggered by specific life situations on a
large scale. A substantial number of participating systems presented
the current state of the art in text classification in general and
transferred it to the task of emotion classification.

Based on the experiences during the organization and preparation of
this shared task, we plan the following steps for a potential second iteration.
The dataset was now constructed via distant supervision, which might be a
cause for inconsistencies in the dataset. We plan to use crowdsourcing
as applied for the estimation of human performance to improve
preprocessing of the data.  In addition, as one participant noted, the
emotion words which were used to retrieve the data were removed,
but, in a subset of the data, other emotion words were retained.

The next step, which we suggest to the participants and future
researchers is introspection of the models -- carefully analyse them
to prove that the models actually learn to infer emotions from subtle
descriptions of situations, instead of purely associating emotion
words with emotion labels. Similarly, an open research question is how
models developed on the IEST data perform on other data
sets. \newcite{Bostan2018} showed that transferring models from one
corpus to another in emotion analysis leads to drops in
performance. Therefore, an interesting option is to use transfer
learning from established corpora (which do not distinguish explicit
and implicit emotion statements) to the IEST data and compare the
models to those directly trained on the IEST and vice versa.

Finally, another line of future research is the application of the
knowledge inferred to other tasks, such as argument mining and
sentiment analysis.

\section*{Acknowledgments}
This work has been partially supported by the German Research Council
(DFG), project SEAT (Structured Multi-Domain Emotion Analysis from
Text, KL 2869/1-1). We thank Evgeny Kim, Laura Bostan, Jeremy Barnes,
and Veronique Hoste for fruitful discussions.

\appendix

\onecolumn

\section{Results by emotion class}

Table \ref{tab:emotresults} shows breakdown of the results by emotion class.\\[\baselineskip]

%cat x | tr '&' '\t' | sed 's/\\//g' | grep -v "^sep" | awk -F'\t' 'NR>0{ for ( i = 2 ; i<=19; i=i+1 ) { $i=sprintf("%.0f",$i);   }}1' OFS="\t"

\newcommand{\sep}{\cmidrule(r){1-1}\cmidrule(rl){2-4}\cmidrule(rl){6-8}\cmidrule(rl){10-12}\cmidrule(rl){14-16}\cmidrule(lr){18-20}\cmidrule(l){22-24}}
\centering\small
\setlength\tabcolsep{5pt}
\begin{tabular}{lccccccccccccccccccccccc}
\toprule
&\multicolumn{3}{c}{Joy}&
&\multicolumn{3}{c}{Sadness}&
&\multicolumn{3}{c}{Disgust}&
&\multicolumn{3}{c}{Anger}&
&\multicolumn{3}{c}{Surprise}&
&\multicolumn{3}{c}{Fear}\\
\sep
Team & P & R & \F && P & R & \F && P & R & \F && P & R & \F && P & R &\F && P & R & \F \\
\sep
Amobee&\textbf{82}&\textbf{82}&\textbf{82}&&70&\textbf{68}&\textbf{69}&&\textbf{73}&70&\textbf{72}&&62&\textbf{66}&\textbf{64}&&\textbf{66}&70&68&&\textbf{77}&73&\textbf{75}\\
IIIDYT&79&81&80&&\textbf{71}&67&69&&70&\textbf{71}&71&&\textbf{66}&63&64&&\textbf{66}&\textbf{71}&\textbf{68}&&76&\textbf{74}&75\\
NTUA-SLP&81&77&79&&71&66&69&&72&70&71&&63&64&63&&62&71&67&&75&73&74\\
\sep
UBC-NLP&79&79&79&&67&67&67&&69&68&69&&62&63&62&&65&67&66&&73&73&73\\
Sentylic&80&77&79&&68&66&67&&69&69&69&&63&61&62&&63&69&66&&73&73&73\\
HUMIR&77&78&78&&70&64&66&&70&68&69&&61&63&62&&61&69&65&&74&70&72\\
nlp&77&78&78&&68&62&65&&70&67&69&&62&63&62&&62&68&65&&72&72&72\\
DataSEARCH&77&77&77&&66&64&65&&69&68&68&&61&62&62&&64&65&65&&72&71&71\\
YNU1510&78&75&76&&64&64&64&&68&68&68&&60&63&62&&64&65&64&&73&71&72\\
EmotiKLUE&77&78&77&&69&59&64&&67&67&67&&60&61&60&&60&68&64&&72&69&71\\
wojtek.pierre&77&75&76&&67&61&64&&66&68&67&&57&60&58&&62&63&62&&69&70&69\\
hgsgnlp&75&75&75&&66&59&62&&67&66&67&&59&59&59&&59&67&63&&69&69&69\\
UWB&74&77&75&&61&68&64&&74&59&65&&57&63&60&&66&56&61&&65&73&69\\
NL-FIIT&76&74&75&&62&64&63&&69&63&66&&61&57&59&&58&65&61&&68&70&69\\
TubOslo&82&67&74&&62&63&62&&62&68&65&&59&56&58&&57&66&62&&68&66&67\\
YNU\_Lab&74&74&74&&66&56&61&&63&67&65&&55&61&58&&63&56&60&&66&70&68\\
Braint&77&70&73&&61&60&60&&60&68&64&&56&55&55&&60&57&59&&63&66&65\\
EmoNLP&73&72&73&&62&57&60&&63&62&63&&55&56&56&&56&61&58&&64&64&64\\
RW&71&72&72&&60&57&59&&62&63&62&&55&52&53&&56&60&58&&62&63&63\\
\sep
Baseline&69&71&70&&58&54&56&&62&62&62&&54&51&52&&55&59&57&&63&63&63\\
\sep
USI-IR&71&69&70&&58&51&54&&59&59&59&&49&58&53&&57&50&53&&59&62&61\\
THU\_NGN&77&78&77&&69&63&66&&68&68&68&&60&63&62&&61&66&64&&71&68&70\\
SINAI&68&68&68&&52&52&52&&59&60&59&&52&51&52&&56&55&55&&61&61&61\\
UTFPR&64&53&58&&54&60&57&&59&58&58&&50&53&52&&51&62&56&&66&56&61\\
CNHZ2017&65&70&67&&58&47&52&&58&59&59&&51&48&50&&49&58&53&&58&57&58\\
lyb3b&72&64&68&&58&46&52&&55&62&58&&46&53&50&&47&50&49&&60&58&59\\
AdobeResearch&62&65&63&&52&52&52&&52&51&52&&48&45&46&&49&52&50&&56&54&55\\
Anonymous&76&77&76&&64&67&65&&70&64&67&&62&59&60&&59&69&64&&74&68&71\\
dinel&61&61&61&&52&37&43&&52&49&50&&44&50&47&&44&54&48&&51&50&50\\
CHANDA&46&64&54&&39&36&38&&54&42&47&&38&37&37&&51&20&29&&39&58&46\\
NLP\_LDW&33&38&36&&18&12&14&&20&31&25&&22&26&24&&18&\phantom{0}7&10&&18&17&18\\
\bottomrule
\end{tabular}
\captionof{table}{\label{tab:emotresults} Results by emotion
  class. Note that this table is limited to the six emotion labels of
  interest in the data set. However, other labels predicted than these
six were taken into account for calculation of the final macro \F
score. Therefore, the macro \F calculated from this table is different
from the results in Table~\ref{tab:offresults} in two cases (THU\_NGN
and Anonymous, who would be on rank 9 and rank 10, when predictions
for classes outside the labels were ignored.).}

\newpage 

\section{Examples}
Table~\ref{tab:allwrongallcorrect} shows examples which have been
correctly or wrongly predicted by all instances. They are discussed in
Section~\ref{sec:erroranalysis}.\\[\baselineskip]

  \centering\footnotesize\sffamily
  \setlength\tabcolsep{1mm}
  \renewcommand*{\arraystretch}{0.8}
  \newcommand{\rtt}[1]{\rotatebox{90}{#1}}
  \begin{tabularx}{\linewidth}{lcX}
    \toprule
    Emo. & $+$/$-$ & Instance \\
    \cmidrule(r){1-1}\cmidrule(rl){2-2}\cmidrule(l){3-3}
    \multirow{11}{*}{\rtt{Anger}} 
    & $+$ & You can't spend your whole life holding the door open for
    people and then being TRIGGER when they dont thank
    you. Nobody asked you to do it. \\
%    & $+$ & I actually get so TRIGGER when I get constant slow
%    replies, it's like if you ain't got time don't text me in the
%    first place \\
    & $+$ & I get impatient and TRIGGER when I'm hungry \\
%    & $+$ & @USERNAME Don't worry about the haters (@USERNAME)... They
%    are just TRIGGER because the TRUTH you speak contradicts
%    the lie they live. \\
    & $+$ & Anyone have the first fast and TRIGGER that I can
    borrow? \\
    \cmidrule(l){2-3}
    & $-$ & I'm kinda TRIGGER that I have to work on Father's
    Day \\
    & $-$ & @USERNAME she'll become TRIGGER that I live close
    by and she will find me and punch me \\
%    & $-$ & @USERNAME Stop that hahahaha, just think about how
%    TRIGGER that woman is right now \\
%    & $-$ & I really feel so TRIGGER when some Disney CM don't
%    like Disney and even hate it, i'm like "PLEASE GIVE ME YOUR JOB"
%    \\
    & $-$ & This has been such a miserable day and I'm TRIGGER
    because I wish I could've enjoyed myself more \\
    \cmidrule(r){1-1}\cmidrule(rl){2-2}\cmidrule(l){3-3}
    \multirow{11}{*}{\rtt{Disgust}} 
%    & $+$ & I feel so TRIGGER when I wear makeup to the gym \\
    & $+$ & I find it TRIGGER when I can see your underwear
    through your leggings \\
    & $+$ & @USERNAME ew ew eeww your weird  I can't I would feel so
    TRIGGER when people touch my hair \\
%    & $+$ & @USERNAME I feel so TRIGGER because of all the
%    sweat even though I had a shower when I got in \\
    & $+$ & nyc smells TRIGGER when it's wet. \\
    \cmidrule(l){2-3}
%    & $-$ & I'm so TRIGGER that I had to block her. \\
    & $-$ & I wanted a cup of coffee for the train ride.  Got ignored
    twice.  I left TRIGGER because I can't afford to miss my
    train. \#needcoffee :( \\
    & $-$ & So this thing where other black people ask where you're
    "really" from then act TRIGGER when you reply with some US
    state. STAHP \\
    & $-$ & I'm so TRIGGER that I have to go to the post
    office to get my jacket that i ordered because delivering it was
    obviously rocket science \\
%    & $-$ & I'm really TRIGGER that there's no taco emoji
%    @USERNAME Apple \\
    \cmidrule(r){1-1}\cmidrule(rl){2-2}\cmidrule(l){3-3}
    \multirow{11}{*}{\rtt{Fear}} 
%    & $+$ &  2 phones calls while on hold, but too TRIGGER
%    that I would hang up on the main call... \\
    & $+$ & @USERNAME \& explain how much the boys mean to me but I'm
    too TRIGGER that they'll just laugh at me bc my dad
    laughed after he \\
    & $+$ & I threw up in a parking lot last night. I'm
    TRIGGER that's becoming my thing. \#illbutmostlymentally \\
%    & $+$ & I'm actually TRIGGER that if I lay down for a
%    nap... I will wake up super stupid at 4:30 am. \\
    & $+$ & When you holding back your emotions and you're
    TRIGGER that when someone tries to comfort you they'll
    come spilling out http://url.removed \\
    \cmidrule(l){2-3}
    & $-$ & It's so funny how people come up to me at work speaking
    Portuguese and they get TRIGGER when I respond in
    Portuguese \\
%    & $-$ & @USERNAME you looked so TRIGGER when i finished
%    lmfao but ilysmmmm(: \\
    & $-$ & @USERNAME it seems so fun but i haven't got to try it
    yet. my mom and sis are always TRIGGER when i try do
    something new with food. \\
    & $-$ & @USERNAME It's hard to be TRIGGER when your giggle
    is so cute \\
%    & $-$ & @USERNAME I was so TRIGGER when I saw it \\
    \cmidrule(r){1-1}\cmidrule(rl){2-2}\cmidrule(l){3-3}
    \multirow{11}{*}{\rtt{Joy}} 
    & $+$ & maybe im so unTRIGGER because i never see the
    sunlight? \\
    & $+$ & @USERNAME you're so welcome !! i'm super TRIGGER
    that i've discovered ur work ! cant wait to see more !! \\
%    & $+$ & @USERNAME So TRIGGER that @USERNAME is back. \\
    & $+$ & @USERNAME Im so TRIGGER that you guys had fun love
    you ♡♡♡♡♡♡ \\
%    & $+$ & @USERNAME I'm so TRIGGER that you found a new
%    friend!!!! Buddy is smiling right now for sure \\
    \cmidrule(l){2-3}
    & $-$ & @USERNAME Not TRIGGER that your show is a
    rerun. It seems every week one or more your segments is a rerun. \\
%    & $-$ & @USERNAME That's most kind.  Your readership and feedback
%    is plenty.  I am not TRIGGER when the numbers go up, only
%    when there's genuine human contact. \\
    & $-$ & I am actually TRIGGER when not invited to certain
    things. I don't have the time and patience to pretend. \\
%    & $-$ & Kinda TRIGGER because I lost 5 pounds in like a 2
%    weeks \\
    & $-$ & This has been such a miserable day and I'm TRIGGER
    because I wish I could've enjoyed myself more \\
    \cmidrule(r){1-1}\cmidrule(rl){2-2}\cmidrule(l){3-3}
    \multirow{11}{*}{\rtt{Sadness}} 
%    & $+$ & No. Sooheart. Sooyoung barely has fansite and it's so
%    TRIGGER that sooheart is closing down. :'( \\
    & $+$ & this award honestly made me so TRIGGER because my
    teacher is leaving  http://url.removed \\
    & $+$ &  It is very TRIGGER that people think depression
    actually does work like that... http://url.removed \\
%    & $+$ & @USERNAME is it TRIGGER that i feel more for nat
%    more than anyone? having to participate in your best friend's
%    suicide is like. jesus. \\
    & $+$ & @USERNAME @USERNAME @USERNAME It's also TRIGGER
    that you so hurt about it :'( \\
    \cmidrule(l){2-3}
    & $-$ & Some bitch stole my seat then I had to steal the seat next
    to me. The boy looked TRIGGER when he saw me, and he was
    smart! \#iwasgonnapass \\
    & $-$ & I was so TRIGGER because I was having fun lol then
    i slipped cus I wasn't wearing shoes \\
%    & $-$ &        Forgot that my cleats were falling apart and got a
%    little TRIGGER when I saw the hole on em \\
    & $-$ & @USERNAME I wipe at my eyes next, then swim a bit. "I'm
    sorry." I repeat, TRIGGER that I made him worry. \\
%    & $-$ & @USERNAME Dolly does not yawn. She is TRIGGER that
%    forget about fun things. \\
    \cmidrule(r){1-1}\cmidrule(rl){2-2}\cmidrule(l){3-3}
    \multirow{11}{*}{\rtt{Surprise}} 
    & $+$ & why am i not TRIGGER that cal said that \\
    & $+$ & @USERNAME why am I not TRIGGER that you're the
    founder \\
    & $+$ & @USERNAME I'm still TRIGGER when students know my
    name. I'm usually just "that guy who wears bow ties" =) (and there
    are a few at WC!) \\
%    & $+$ & @USERNAME I know! Not TRIGGER when you consider
%    more Americans watch football than vote. As a football fan, all I
%    can say is Go Hawks! \\
%    & $+$ & @USERNAME @USERNAME Not TRIGGER when you consider his background. \\
    \cmidrule(l){2-3}
    & $-$ & It's TRIGGER when I see people that have the same
    phone as me no has htcs \\
    & $-$ & There is a little boy in here who is TRIGGER that
    he has to pay for things and that we won't just give him things \\
    & $-$ & totally TRIGGER that my fams celebrating easter
    today  because my sister goes back to uni sunday \\
%    & $-$ & My cool brother is actually impressed by me and is proud
%    of everything, and my other brother is really salty and
%    TRIGGER that I hate him \\
%    & $-$ & I am so TRIGGER that our veterans are being
%    treated this way! http://url.removed \\
    \bottomrule
  \end{tabularx}
  \captionof{table}{\label{tab:allwrongallcorrect} Subsample of Tweets that were correctly predicted by all teams and of Tweets that were not correctly predicted by any team.}

\end{document}